# Data Fusion by Matrix Factorization

**Marinka Žitnik**                                                                          MARINKA.ZITNIK@FRI.UNI-LJ.SI
Faculty of Computer and Information Science, University of Ljubljana, Tržaška 25, SI-1000 Ljubljana, Slovenia

**Blaž Zupan**                                                                              BLAZ.ZUPAN@FRI.UNI-LJ.SI
Faculty of Computer and Information Science, University of Ljubljana, Tržaška 25, SI-1000 Ljubljana, Slovenia
Department of Molecular and Human Genetics, Baylor College of Medicine, Houston, TX-77030, USA

## Abstract

For most problems in science and engineering we can obtain data that describe the system from various perspectives and record the behaviour of its individual components. Heterogeneous data sources can be collectively mined by data fusion. Fusion can focus on a specific target relation and exploit directly associated data together with data on the context or additional constraints. In the paper we describe a data fusion approach with penalized matrix tri-factorization that simultaneously factorizes data matrices to reveal hidden associations. The approach can directly consider any data sets that can be expressed in a matrix, including those from attribute-based representations, ontologies, associations and networks. We demonstrate its utility on a gene function prediction problem in a case study with eleven different data sources. Our fusion algorithm compares favourably to state-of-the-art multiple kernel learning and achieves higher accuracy than can be obtained from any single data source alone.

## 1. Introduction

Data abounds in all areas of human endeavour. Big data (Cuzzocrea et al., 2011) is not only large in volume and may include thousands of features, but it is also heterogeneous. We may gather various data sets that are directly related to the problem, or data sets that are loosely related to our study but could be useful when combined with other data sets. Consider, for example, the exposome (Rappaport & Smith, 2010) that encompasses the totality of human endeav-our in the study of disease. Let us say that we examine susceptibility to a particular disease and have access to the patients' clinical data together with data on their demographics, habits, living environments, friends, relatives, and movie-watching habits and genre ontology. Mining such a diverse data collection may reveal interesting patterns that would remain hidden if considered only directly related, clinical data. What if the disease was less common in living areas with more open spaces or in environments where people need to walk instead of drive to the nearest grocery? Is the disease less common among those that watch comedies and ignore politics and news?

Methods for data fusion can collectively treat data sets and combine diverse data sources even when they differ in their conceptual, contextual and typographical representation (Aerts et al., 2006; Boström et al., 2007). Individual data sets may be incomplete, yet because of their diversity and complementarity, fusion improves the robustness and predictive performance of the resulting models.

Data fusion approaches can be classified into three main categories according to the modeling stage at which fusion takes place (Pavlidis et al., 2002; Schölkopf et al., 2004; Maragos et al., 2008; Greene & Cunningham, 2009). *Early (or full) integration* transforms all data sources into a single, feature-based table and treats this as a single data set. This is theoretically the most powerful scheme of multimodal fusion because the inferred model can contain any type of relationships between the features from within and between the data sources. Early integration relies on procedures for feature construction. For our exposome example, patient-specific data would need to include both clinical data and information from the movie genre ontologies. The former may be trivial as this data is already related to each specific patient, while the latter requires more complex feature engineering.

Early integration neglects the modular structure of the data.

In *late (decision) integration*, each data source gives rise to a separate model. Predictions of these models are fused by model weighting. Again, prior to model inference, it is necessary to transform each data set to encode relations to the target concept. For our example, information on the movie preferences of friends and relatives would need to be mapped to disease associations. Such transformations may not be trivial and would need to be crafted independently for each data source. Once the data are transformed, the fusion can utilize any of the existing ensemble methods.

The youngest branch of data fusion algorithms is *intermediate (partial) integration*. It relies on algorithms that explicitly address the multiplicity of data and fuse them through inference of a joint model. Intermediate integration does not fuse the input data, nor does it develop separate models for each data source. It instead retains the structure of the data sources and merges them at the level of a predictive model. This particular approach is often preferred because of its superior predictive accuracy (Pavlidis et al., 2002; Lanckriet et al., 2004b; Gevaert et al., 2006; Tang et al., 2009; van Vliet et al., 2012), but for a given model type, it requires the development of a new inference algorithm.

In this paper we report on the development of a new method for intermediate data fusion based on constrained matrix factorization. Our aim was to construct an algorithm that requires only minimal transformation (if any at all) of input data and can fuse attribute-based representations, ontologies, associations and networks. We begin with a description of related work and the background of matrix factorization. We then present our data fusion algorithm and finally demonstrate its utility in a study comparing it to state-of-the-art intermediate integration with multiple kernel learning and early integration with random forests.

## 2. Background and Related Work

Approximate matrix factorization estimates a data matrix $\mathbf{R}$ as a product of low-rank matrix factors that are found by solving an optimization problem. In two-factor decomposition, $\mathbf{R} = \mathbb{R}^{n \times m}$ is decomposed to a product $\mathbf{WH}$, where $\mathbf{W} = \mathbb{R}^{n \times k}$, $\mathbf{H} = \mathbb{R}^{k \times m}$ and $k \ll \min(n, m)$. A large class of matrix factorization algorithms minimize discrepancy between the observed matrix and its low-rank approximation, such that $\mathbf{R} \approx \mathbf{WH}$. For instance, SVD, non-negative matrix factorization and exponential family PCA all min-imize Bregman divergence (Singh & Gordon, 2008). The cost function can be further extended with various constraints (Zhang et al., 2011; Wang et al., 2012). Wang *et al.* (2008) (Wang et al., 2008) devised a penalized matrix tri-factorization to include a set of must-link and cannot-link constraints. We exploit this approach to include relations between objects of the same type. Constraints would for instance allow us to include information from movie genre ontologies or social network friendships.

Although often used in data analysis for dimensionality reduction, clustering or low-rank approximation, there have been few applications of matrix factorization for data fusion. Lange *et al.* (2005) (Lange & Buhmann, 2005) proposed an early integration by non-negative matrix factorization of a target matrix, which was a convex combination of similarity matrices obtained from multiple information sources. Their work is similar to that of Wang *et al.* (2012) (Wang et al., 2012), who applied non-negative matrix tri-factorization with input matrix completion.

Zhang *et al.* (2012) (Zhang et al., 2012) proposed a joint matrix factorization to decompose a number of data matrices $\mathbf{R}_i$ into a common basis matrix $\mathbf{W}$ and different coefficient matrices $\mathbf{H}_i$, such that $\mathbf{R}_i \approx \mathbf{WH}_i$ by minimizing $\sum_i ||\mathbf{R}_i - \mathbf{WH}_i||$. This is an intermediate integration approach with different data sources that describe objects of the same type. A similar approach but with added network-regularized constraints has also been proposed (Zhang et al., 2011). Our work extends these two approaches by simultaneously dealing with heterogeneous data sets and objects of different types.

In the paper we use a variant of three-factor matrix factorization that decomposes $\mathbf{R}$ into $\mathbf{G} \in \mathbb{R}^{n \times k_1}$, $\mathbf{F} \in \mathbb{R}^{m \times k_2}$ and $\mathbf{S} \in \mathbb{R}^{k_1 \times k_2}$ such that $\mathbf{R} \approx \mathbf{GSF}^T$ (Wang et al., 2008). Approximation can be rewritten such that entry $\mathbf{R}(p, q)$ is approximated by an inner product of the $p$-th row of matrix $\mathbf{G}$ and a linear combination of the columns of matrix $\mathbf{S}$, weighted by the $q$-th column of $\mathbf{F}$. The matrix $\mathbf{S}$, which has relatively few vectors compared to $\mathbf{R}$ ($k_1 \ll n$, $k_2 \ll m$), is used to represent many data vectors, and a good approximation can only be achieved in the presence of the latent structure in the original data.

We are currently witnessing increasing interest in the joint treatment of heterogeneous data sets and the emergence of approaches specifically designed for data fusion. These include canonical correlation analysis (Chaudhuri et al., 2009), combining many interaction networks into a composite network (Mostafavi & Morris, 2012), multiple graph clustering with linked matrix factorization (Tang et al., 2009), a mixture of Markov chains associated with different graphs (Zhou & Burges, 2007), dependency-seeking clustering algorithms with variational Bayes (Klami & Kaski, 2008), latent factor analysis (Lopes et al., 2011; Luttinen & Ilin, 2009), nonparametric Bayes ensemble learning (Xing & Dunson, 2011), approaches based on Bayesian theory (Zhang & Ji, 2006; Alexeyenko & Sonnhammer, 2009; Huttenhower et al., 2009), neural networks (Carpenter et al., 2005) and module guided random forests (Chen & Zhang, 2013). These approaches either fuse input data (early integration) or predictions (late integration) and do not directly combine heterogeneous representation of objects of different types.

A state-of-the-art approach to intermediate data integration is kernel-based learning. Multiple kernel learning (MKL) has been pioneered by Lanckriet *et al.* (2004) (Lanckriet et al., 2004a) and is an additive extension of single kernel SVM to incorporate multiple kernels in classification, regression and clustering. The MKL assumes that $\mathcal{E}_1, \ldots, \mathcal{E}_r$ are $r$ different representations of the same set of $n$ objects. Extension from single to multiple data sources is achieved by additive combination of kernel matrices, given by $\Omega = \left\{ \sum_{i=1}^{r} \theta_i \mathbf{K}_i \mid \forall i : \theta_i \geq 0, \sum_{i=1}^{r} \theta_i^\delta = 1, \mathbf{K}_i \succeq 0 \right\}$, where $\theta_i$ are weights of the kernel matrices, $\delta$ is a parameter determining the norm of constraint posed on coefficients (for $L_2, L_p$-norm MKL, see (Kloft et al., 2009; Yu et al., 2010; 2012)) and $\mathbf{K}_i$ are normalized kernel matrices centered in the Hilbert space. Among other improvements, Yu *et al.* (2010) extended the framework of the MKL in Lanckriet *et al.* (2004) (Lanckriet et al., 2004a) by optimizing various norms in the dual problem of SVMs that allows non-sparse optimal kernel coefficients $\theta_i^*$. The heterogeneity of data sources in the MKL is resolved by transforming different object types and data structures (e.g., strings, vectors, graphs) into kernel matrices. These transformations depend on the choice of the kernels, which in turn affect the method's performance (Debnath & Takahashi, 2004).

## 3. Data Fusion Algorithm

Our data fusion algorithm considers $r$ object types $\mathcal{E}_1, \ldots, \mathcal{E}_r$ and a collection of data sources, each relating a pair of object types $(\mathcal{E}_i, \mathcal{E}_j)$. In our introductory example of the exposome, object types could be a patient, a disease or a living environment, among others. If there are $n_i$ objects of type $\mathcal{E}_i$ ($o_p^i$ is $p$-th object of type $\mathcal{E}_i$) and $n_j$ objects of type $\mathcal{E}_j$, we represent the observations from the data source that relates $(\mathcal{E}_i, \mathcal{E}_j)$ for $i \neq j$ in a sparse matrix $\mathbf{R}_{ij} \in \mathbb{R}^{n_i \times n_j}$. An example of such a matrix would be one that relates patients and drugs by reporting on their current prescriptions. Notice that in general matrices $\mathbf{R}_{ij}$ and $\mathbf{R}_{ji}$ need not be symmetric. A data source that provides relations between objects of the same type $\mathcal{E}_i$ is represented by a constraint matrix $\mathbf{\Theta}_i \in \mathbb{R}^{n_i \times n_i}$. Examples of such constraints are social networks and drug interactions. In real-world scenarios we might not have access to relations between all pairs of object types. Our data fusion algorithm still integrates all available data if an underlying graph of relations between object types is connected. Fig. 1 shows an example of a block configuration of the fusion system with four object types.

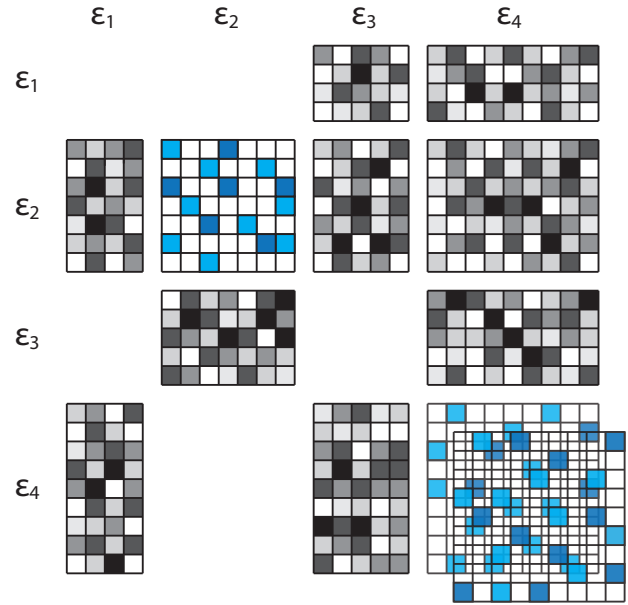

Figure 1. Conceptual fusion configuration of four object types, $\mathcal{E}_1, \mathcal{E}_2, \mathcal{E}_3$ and $\mathcal{E}_4$. Data sources relate pairs of object types (matrices with shades of gray). For example, data matrix $\mathbf{R}_{42}$ relates object types $\mathcal{E}_4$ and $\mathcal{E}_2$. Some relations are missing; there is no data source relating $\mathcal{E}_3$ and $\mathcal{E}_1$. Constraint matrices (in blue) relate objects of the same type. In our example, constraints are provided for object types $\mathcal{E}_2$ (one constraint matrix) and $\mathcal{E}_4$ (two constraint matrices).

To retain the block structure of our system, we propose the simultaneous factorization of all relation matrices $\mathbf{R}_{ij}$ constrained by $\mathbf{\Theta}_i$. The resulting system contains factors that are specific to each data source and factors that are specific to each object type. Through factor sharing we fuse the data but also identify source-specific patterns.

The proposed fusion approach is different from treating an entire system (e.g., from Fig. 1) as a large single matrix. Factorization of such a matrix would yield factors that are not object type-specific and would thus disregard the structure of the system. We also show (Sec. 5.5) that such an approach is inferior in terms of predictive performance.

We apply data fusion to infer relations between two target object types, $\mathcal{E}_i$ and $\mathcal{E}_j$ (Sec. 3.4 and Sec. 3.5). This relation, encoded in a target matrix $\mathbf{R}_{ij}$, will be observed in the context of all other data sources (Sec. 3.1). We assume that $\mathbf{R}_{ij}$ is a $[0,1]$-matrix that is only partially observed. Its entries indicate a degree of relation, 0 denoting no relation and 1 denoting the strongest relation. We aim to predict unobserved entries in $\mathbf{R}_{ij}$ by reconstructing them through matrix factorization. Such treatment in general applies to multi-class or multi-label classification tasks, which are conveniently addressed by multiple kernel fusion (Yu et al., 2010), with which we compare our performance in this paper.

### 3.1. Factorization

An input to data fusion is a relation block matrix $\mathbf{R}$ that conceptually represents all relation matrices:

$$\mathbf{R} = \begin{bmatrix} \mathbf{0} & \mathbf{R}_{12} & \cdots & \mathbf{R}_{1r} \\ \mathbf{R}_{21} & \mathbf{0} & \cdots & \mathbf{R}_{2r} \\ \vdots & \vdots & \ddots & \vdots \\ \mathbf{R}_{r1} & \mathbf{R}_{r2} & \cdots & \mathbf{0} \end{bmatrix}. \quad (1)$$

A block in the $i$-th row and $j$-th column ($\mathbf{R}_{ij}$) of matrix $\mathbf{R}$ represents the relationship between object type $\mathcal{E}_i$ and $\mathcal{E}_j$. The $p$-th object of type $\mathcal{E}_i$ (i.e. $o_p^i$) and $q$-th object of type $\mathcal{E}_j$ (i.e. $o_q^j$) are related by $\mathbf{R}_{ij}(p, q)$.

We additionally consider constraints relating objects of the same type. Several data sources may be available for each object type. For instance, personal relations may be observed from a social network or a family tree. Assume there are $t_i \geq 0$ data sources for object type $\mathcal{E}_i$ represented by a set of constraint matrices $\mathbf{\Theta}_i^{(t)}$ for $t \in \{1, 2, \ldots, t_i\}$. Constraints are collectively encoded in a set of constraint block diagonal matrices $\mathbf{\Theta}^{(t)}$ for $t \in \{1, 2, \ldots, \max_i t_i\}$:

$$\mathbf{\Theta}^{(t)} = \begin{bmatrix} \mathbf{\Theta}_1^{(t)} & \mathbf{0} & \cdots & \mathbf{0} \\ \mathbf{0} & \mathbf{\Theta}_2^{(t)} & \cdots & \mathbf{0} \\ \vdots & \vdots & \ddots & \vdots \\ \mathbf{0} & \mathbf{0} & \cdots & \mathbf{\Theta}_r^{(t)} \end{bmatrix}. \quad (2)$$

The $i$-th block along the main diagonal of $\mathbf{\Theta}^{(t)}$ is zero if $t > t_i$. Entries in constraint matrices are positive for objects that are not similar and negative for objects that are similar. The former are known as cannot-link constraints because they impose penalties on the current approximation of the matrix factors, and the latter are must-link constraints, which are rewards that reduce the value of the cost function during optimization.

The block matrix $\mathbf{R}$ is tri-factorized into block matrix factors $\mathbf{G}$ and $\mathbf{S}$:

$$\mathbf{G} = \begin{bmatrix} \mathbf{G}_1^{n_1 \times k_1} & \mathbf{0} & \cdots & \mathbf{0} \\ \mathbf{0} & \mathbf{G}_2^{n_2 \times k_2} & \cdots & \mathbf{0} \\ \vdots & \vdots & \ddots & \vdots \\ \mathbf{0} & \mathbf{0} & \cdots & \mathbf{G}_r^{n_r \times k_r} \end{bmatrix},$$

$$\mathbf{S} = \begin{bmatrix} \mathbf{0} & \mathbf{S}_{12}^{k_1 \times k_2} & \cdots & \mathbf{S}_{1r}^{k_1 \times k_r} \\ \mathbf{S}_{21}^{k_2 \times k_1} & \mathbf{0} & \cdots & \mathbf{S}_{2r}^{k_2 \times k_r} \\ \vdots & \vdots & \ddots & \vdots \\ \mathbf{S}_{r1}^{k_r \times k_1} & \mathbf{S}_{r2}^{k_r \times k_2} & \cdots & \mathbf{0} \end{bmatrix}. \quad (3)$$

A factorization rank $k_i$ is assigned to $\mathcal{E}_i$ during inference of the factorized system. Factors $\mathbf{S}_{ij}$ define the relation between object types $\mathcal{E}_i$ and $\mathcal{E}_j$, while factors $\mathbf{G}_i$ are specific to objects of type $\mathcal{E}_i$ and are used in the reconstruction of every relation with this object type. In this way, each relation matrix $\mathbf{R}_{ij}$ obtains its own factorization $\mathbf{G}_i \mathbf{S}_{ij} \mathbf{G}_j^T$ with factor $\mathbf{G}_i$ ($\mathbf{G}_j$) that is shared across relations which involve object types $\mathcal{E}_i$ ($\mathcal{E}_j$). This can also be observed from the block structure of the reconstructed system $\mathbf{GSG}^T$:

$$\begin{bmatrix} \mathbf{0} & \mathbf{G}_1 \mathbf{S}_{12} \mathbf{G}_2^T & \cdots & \mathbf{G}_1 \mathbf{S}_{1r} \mathbf{G}_r^T \\ \mathbf{G}_2 \mathbf{S}_{21} \mathbf{G}_1^T & \mathbf{0} & \cdots & \mathbf{G}_2 \mathbf{S}_{2r} \mathbf{G}_r^T \\ \vdots & \vdots & \ddots & \vdots \\ \mathbf{G}_r \mathbf{S}_{r1} \mathbf{G}_1^T & \mathbf{G}_r \mathbf{S}_{r2} \mathbf{G}_2^T & \cdots & \mathbf{0} \end{bmatrix}. \quad (4)$$

The objective function minimized by penalized matrix tri-factorization ensures good approximation of the input data and adherence to must-link and cannot-link constraints. We extend it to include multiple constraint matrices for each object type:

$$\min_{\mathbf{G} \geq 0} ||\mathbf{R} - \mathbf{GSG}^T|| + \sum_{t=1}^{\max_i t_i} \text{tr}(\mathbf{G}^T \mathbf{\Theta}^{(t)} \mathbf{G}), \quad (5)$$

where $|| \cdot ||$ and $\text{tr}(\cdot)$ denote the Frobenius norm and trace, respectively.

For decomposition of relation matrices we derive updating rules based on the penalized matrix tri-factorization of Wang et al. (2008) (Wang et al., 2008).

The algorithm for solving the optimization problem in Eq. (5) initializes matrix factors (see Sec. 3.6) and improves them iteratively. Successive updates of $\mathbf{G}_i$ and $\mathbf{S}_{ij}$ converge to a local minimum of the optimization problem.

Multiplicative updating rules are derived from Wang *et al.* (2008) (Wang et al., 2008) by fixing one matrix factor (i.e. $\mathbf{G}$) and considering the roots of the partial derivative with respect to the other matrix factor (i.e. $\mathbf{S}$, and vice-versa) of the Lagrangian function. The latter is constructed from the objective function given in Eq. (5). The update rule alternates between first fixing $\mathbf{G}$ and updating $\mathbf{S}$, and then fixing $\mathbf{S}$ and updating $\mathbf{G}$, until convergence. The update rule for $\mathbf{S}$ is:

$$\mathbf{S} \leftarrow (\mathbf{G}^T\mathbf{G})^{-1}\mathbf{G}^T\mathbf{R}\mathbf{G}(\mathbf{G}^T\mathbf{G})^{-1} \quad (6)$$

The rule for factor $\mathbf{G}$ incorporates data on constraints. We update each element of $\mathbf{G}$ at position $(p, q)$ by multiplying it with:

$$\sqrt{\frac{(\mathbf{RGS})^+_{(p,q)} + [\mathbf{G}(\mathbf{SG}^T\mathbf{GS})^-]_{(p,q)} + \sum_t ((\mathbf{\Theta}^{(t)})^- \mathbf{G})_{(p,q)}}{(\mathbf{RGS})^-_{(p,q)} + [\mathbf{G}(\mathbf{SG}^T\mathbf{GS})^+]_{(p,q)} + \sum_t ((\mathbf{\Theta}^{(t)})^+ \mathbf{G})_{(p,q)}}}. \quad (7)$$

Here, $\mathbf{X}^+_{(p,q)}$ is defined as $\mathbf{X}(p,q)$ if entry $\mathbf{X}(p,q) \geq 0$ else 0. Similarly, the $\mathbf{X}^-_{(p,q)}$ is $-\mathbf{X}(p,q)$ if $\mathbf{X}(p,q) \leq 0$ else it is set to 0. Therefore, both $\mathbf{X}^+$ and $\mathbf{X}^-$ are non-negative matrices. This definition is applied to matrices in Eq. (7), i.e. $\mathbf{RGS}$, $\mathbf{SG}^T\mathbf{GS}$, and $\mathbf{\Theta}^{(t)}$ for $t \in \{1, 2, \dots, \max_i t_i\}$.

### 3.2. Stopping Criteria

In this paper we apply data fusion to infer relations between two target object types, $\mathcal{E}_i$ and $\mathcal{E}_j$. We hence define the stopping criteria that observes convergence in approximating only the target matrix $\mathbf{R}_{ij}$. Our convergence criteria is:

$$||\mathbf{R}_{ij} - \mathbf{G}_i\mathbf{S}_{ij}\mathbf{G}_j^T|| < \epsilon, \quad (8)$$

where $\epsilon$ is a user-defined parameter, possibly refined through observing log entries of several runs of the factorization algorithm. In our experiments $\epsilon$ was set to $10^{-5}$. To reduce the computational load, the convergence criteria was assessed only every fifth iteration only.

### 3.3. Parameter Estimation

To fuse data sources on $r$ object types, it is necessary to estimate $r$ factorization ranks, $k_1, k_2, \dots, k_r$. These are chosen from a predefined interval of possible values for each rank by estimating the model quality. To reduce the number of needed factorization runs we mimic the bisection method by first testing rank values at the midpoint and borders of specified ranges and then for each rank value selecting the subinterval for which better model quality was achieved. We evaluate the models through the explained variance, the residual sum of squares (RSS) and a measure based on the cophenetic correlation coefficient $\rho$. We compute these measures from the target relation matrix $\mathbf{R}_{ij}$. The RSS is computed from observed entries in $\mathbf{R}_{ij}$ as $\text{RSS}(\mathbf{R}_{ij}) = \sum_{(o^i_p, o^j_q) \in \mathcal{A}(\mathcal{E}_i, \mathcal{E}_j)} \left[ (\mathbf{R}_{ij} - \mathbf{G}_i\mathbf{S}_{ij}\mathbf{G}_j^T)(p, q) \right]^2$, where $\mathcal{A}(\mathcal{E}_i, \mathcal{E}_j)$ is the set of known associations between objects of $\mathcal{E}_i$ and $\mathcal{E}_j$. The explained variance for $\mathbf{R}_{ij}$ is $r^2(\mathbf{R}_{ij}) = 1 - \text{RSS}(\mathbf{R}_{ij})/\sum_{p,q}[\mathbf{R}_{ij}(p, q)]^2$. The cophenetic correlation score was implemented as described in (Brunet et al., 2004).

We assess the three quality scores through internal cross-validation and observe how $r^2(\mathbf{R}_{ij})$, $\text{RSS}(\mathbf{R}_{ij})$ and $\rho(\mathbf{R}_{ij})$ vary as factorization ranks change. We select ranks $k_1, k_2, \dots, k_r$ where the cophenetic coefficient begins to fall, the explained variance is high and the RSS curve shows an inflection point (Hutchins et al., 2008).

### 3.4. Prediction from Matrix Factors

The approximate relation matrix $\widehat{\mathbf{R}}_{ij}$ for the target pair of object types $\mathcal{E}_i$ and $\mathcal{E}_j$ is reconstructed as:

$$\widehat{\mathbf{R}}_{ij} = \mathbf{G}_i\mathbf{S}_{ij}\mathbf{G}_j^T. \quad (9)$$

When the model is requested to propose relations for a new object $o^i_{n_i+1}$ of type $\mathcal{E}_i$ that was not included in the training data, we need to estimate its factorized representation and use the resulting factors for prediction. We formulate a non-negative linear least-squares (NNLS) and solve it with an efficient interior point Newton-like method (Van Benthem & Keenan, 2004) for $\min_{\mathbf{x} \geq 0} ||(\mathbf{SG}^T)^T\mathbf{x} - \mathbf{o}^i_{n_i+1}||_2$, where $\mathbf{o}^i_{n_i+1} \in \mathbb{R}^{\sum n_i}$ is the original description of object $o^i_{n_i+1}$ across all available relation matrices and $\mathbf{x} \in \mathbb{R}^{\sum k_i}$ is its factorized representation. A solution vector $\mathbf{x}^T$ is added to $\mathbf{G}$ and a new $\widehat{\mathbf{R}}_{ij} \in \mathbb{R}^{(n_i+1) \times n_j}$ is computed using Eq. (9).

We would like to identify object pairs $(o^i_p, o^j_q)$ for which the predicted degree of relation $\widehat{\mathbf{R}}_{ij}(p, q)$ is unusually high. We are interested in candidate pairs $(o^i_p, o^j_q)$ for which the estimated association score $\widehat{\mathbf{R}}_{ij}(p, q)$ is greater than the mean estimated score of all known relations of $o^i_p$:

$$\widehat{\mathbf{R}}_{ij}(p, q) > \frac{1}{|\mathcal{A}(o^i_p, \mathcal{E}_j)|} \sum_{o^j_m \in \mathcal{A}(o^i_p, \mathcal{E}_j)} \widehat{\mathbf{R}}_{ij}(p, m), \quad (10)$$

where $\mathcal{A}(o^i_p, \mathcal{E}_j)$ is the set of all objects of $\mathcal{E}_j$ related to $o^i_p$. Notice that this rule is row-centric, that is, given an object of type $\mathcal{E}_i$, it searches for objects of the other type ($\mathcal{E}_j$) that it could be related to. We can modify the rule to become column-centric, or even combine the two rules.

For example, let us consider that we are studying disease predispositions for a set of patients. Let the patients be objects of type $\mathcal{E}_i$ and diseases objects of type $\mathcal{E}_j$. A patient-centric rule would consider a patient and his medical history and through Eq. (10) propose a set of new disease associations. A disease-centric rule would instead consider all patients already associated with a specific disease and identify other patients with a sufficiently high association score.

In our experiments we combine row-centric and column-centric approaches. We first apply a row-centric approach to identify candidates of type $\mathcal{E}_i$ and then estimate the strength of association to a specific object $o_q^j$ by reporting a percentile of association score in the distribution of scores for all true associations of $o_q^j$, that is, by considering the scores in the $q$-ed column of $\widehat{\mathbf{R}}_{ij}$.

### 3.5. An Ensemble Approach to Prediction

Instead of a single model, we construct an ensemble of factorization models. The resulting matrix factors in each model differ due to the initial random conditions or small random perturbations of selected factorization ranks. We use each factorization system for inference of associations (Sec. 3.4) and then select the candidate pair through a majority vote. That is, the rule from Eq. (10) must apply in more than one half of factorized systems of the ensemble. Ensembles improved the predictive accuracy and stability of the factorized system and the robustness of the results. In our experiments the ensembles combined 15 factorization models.

### 3.6. Matrix Factor Initialization

The inference of the factorized system in Sec. 3.1 is sensitive to the initialization of factor $\mathbf{G}$. Proper initialization sidesteps the issue of local convergence and reduces the number of iterations needed to obtain matrix factors of equal quality. We initialize $\mathbf{G}$ by separately initializing each $\mathbf{G}_i$, using algorithms for single-matrix factorization. Factors $\mathbf{S}$ are computed from $\mathbf{G}$ (Eq. (6)) and do not require initialization.

Wang *et al.* (2008) (Wang et al., 2008) and several other authors (Lee & Seung, 2000) use simple random initialization. Other more informed initialization algorithms include random C (Albright et al., 2006), random Acol (Albright et al., 2006), non-negative double SVD and its variants (Boutsidis & Gallopoulos, 2008), and $k$-means clustering or relaxed SVD-centroid initialization (Albright et al., 2006). We show that the latter approaches are indeed better over a random initialization (Sec. 5.4). We use random Acol in our case study. Random Acol computes each column of $\mathbf{G}_i$ as an element-wise average of a random subset of columns in $\mathbf{R}_{ij}$.

## 4. Experiments

We considered a gene function prediction problem from molecular biology, where recent technological advancements have allowed researchers to collect large and diverse experimental data sets. Data integration is an important aspect of bioinformatics and the subject of much recent research (Yu et al., 2010; Vaske et al., 2010; Moreau & Tranchevent, 2012; Mostafavi & Morris, 2012). It is expected that various data sources are complementary and that high accuracy of predictions can be achieved through data fusion (Parikh & Polikar, 2007; Pandey et al., 2010; Savage et al., 2010; Xing & Dunson, 2011). We studied the fusion of eleven different data sources to predict gene function in the social amoeba *Dictyostelium discoideum* and report on the cross-validated accuracy for 148 gene annotation terms (classes).

We compare our data fusion algorithm to an early integration by a random forest (Boulesteix et al., 2008) and an intermediate integration by multiple kernel learning (MKL) (Yu et al., 2010). Kernel-based fusion used a multi-class $L_2$ norm MKL with Vapnik's SVM (Ye et al., 2008). The MKL was formulated as a second order cone program (SOCP) and its dual problem was solved by the conic optimization solver Se-DuMi[1]. Random forests from the Orange[2] data mining suite were used with default parameters.

### 4.1. Data

The social amoeba *D. discoideum*, a popular model organism in biomedical research, has about 12,000 genes, some of which are functionally annotated with terms from Gene Ontology[3] (GO). The annotation is rather sparse and only ∼1,400 genes have experimentally derived annotations. The *Dictyostelium* community can thus gain from computational models with accurate function predictions.

We observed six object types: genes (type 1), ontology terms (type 2), experimental conditions (type 3), publications from the PubMed database (PMID) (type 4), Medical Subject Headings (MeSH) descriptors (type 5), and KEGG[4] pathways (type 6). Object types and their observed relations are shown in Fig. 2. The data included gene expression measured during different time-points of a 24-hour development cycle (Parikh et al., 2010) ($\mathbf{R}_{13}$, 14 experimental conditions), gene annotations with experimental evidence code to 148 generic slim terms from the GO ($\mathbf{R}_{12}$), PMIDs and their associated *D. discoideum* genes from Dictybase[5] ($\mathbf{R}_{14}$), genes participating in KEGG pathways ($\mathbf{R}_{16}$), assignments of MeSH descriptors to publications from PubMed ($\mathbf{R}_{45}$), references to published work on associations between a specific GO term and gene product ($\mathbf{R}_{42}$), and associations of enzymes involved in KEGG pathways and related to GO terms ($\mathbf{R}_{62}$).

---

[1] http://sedumi.ie.lehigh.edu
[2] http://orange.biolab.si
[3] http://www.geneontology.org

To balance $\mathbf{R}_{12}$, our target relation matrix, we added an equal number of non-associations for which there is no evidence of any type in the GO. We constrain our system by considering gene interaction scores from STRING v9.0[6] ($\mathbf{\Theta}_1$) and slim term similarity scores ($\mathbf{\Theta}_2$) computed as $0.8^{\text{hops}}$, where hops is the length of the shortest path between two terms in the GO graph. Similarly, MeSH descriptors are constrained with the average number of hops in the MeSH hierarchy between each pair of descriptors ($\mathbf{\Theta}_5$). Constraints between KEGG pathways correspond to the number of common ortholog groups ($\mathbf{\Theta}_6$). The slim subset of GO terms was used to limit the optimization complexity of the MKL and the number of variables in the SOCP, and to ease the computational burden of early integration by random forests, which inferred a separate model for each of the terms.

To study the effects of data sparseness, we conducted an experiment in which we selected either the 100 or 1000 most GO-annotated genes (see second column of Table 1 for sparsity).

In a separate experiment we examined predictions of gene association with any of nine GO terms that are of specific relevance to the current research in the *Dictyostelium* community (upon consultations with Gad Shaulsky, Baylor College of Medicine, Houston, TX; see Table 2). Instead of using a generic slim subset of terms, we examined the predictions in the context of a complete set of GO terms. This resulted in a data set with ∼2.000 terms, each term having on average 9.64 direct gene annotations.

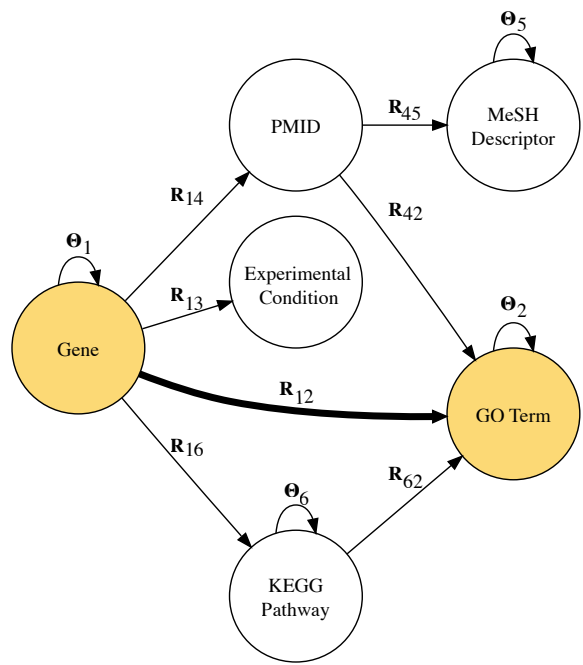

*Figure 2.* The fusion configuration. Nodes represent object types used in our study. Edges correspond to relation and constraint matrices. The arc that represents the target matrix $\mathbf{R}_{12}$ and its object types are highlighted.

### 4.2. Scoring

We estimated the quality of inferred models by tenfold cross-validation. In each iteration, we split the gene set to a train and test set. The data on genes from the test set was entirely omitted from the training data. We developed prediction models from the training data and tested them on the genes from the test set. The performance was evaluated using an $F_1$ score, a harmonic mean of precision and recall, which was averaged across cross-validation runs.

### 4.3. Preprocessing for Kernel-Based Fusion

We generated an RBF kernel for gene expression measurements from $\mathbf{R}_{13}$ with width $\sigma = 0.5$ and the RBF function $\kappa(\mathbf{x}_i, \mathbf{x}_j) = \exp(-||\mathbf{x}_i - \mathbf{x}_j||^2/2\sigma^2)$, and a linear kernel for $[0, 1]$-protein-interaction matrix from $\mathbf{\Theta}_1$. Kernels were applied to data matrices. We used a linear kernel to generate a kernel matrix from *D. discoideum* specific genes that participate in pathways ($\mathbf{R}_{16}$), and a kernel matrix from PMIDs and their associated genes ($\mathbf{R}_{14}$). Several data sources describe relations between object types other than genes. For kernel-based fusion we had to transform them to explicitly relate to genes. For instance, to relate genes and MeSH descriptors, we counted the number of publications that were associated with a specific gene ($\mathbf{R}_{14}$) and were assigned a specific MeSH descriptor ($\mathbf{R}_{45}$, see also Fig. 2). A linear kernel was applied to the resulting matrix. Kernel matrices that incorporated relations between KEGG pathways and GO terms ($\mathbf{R}_{62}$), and publications and GO terms were obtained in similar fashion.

To represent the hierarchical structure of MeSH descriptors ($\mathbf{\Theta}_5$), the semantic structure of the GO graph ($\mathbf{\Theta}_2$) and ortholog groups that correspond to KEGG pathways ($\mathbf{\Theta}_6$), we considered the genes as nodes in three distinct large weighted graphs. In the graph for $\mathbf{\Theta}_5$, the link between two genes was weighted by the similarity of their associated sets of MeSH descriptors using information from $\mathbf{R}_{14}$ and $\mathbf{R}_{45}$. We considered the MeSH hierarchy to measure these similarities. Similarly, for the graph for $\mathbf{\Theta}_2$ we considered the GO semantic structure in computing similarities of sets of GO terms associated with genes. In the graph for $\mathbf{\Theta}_6$, the gene edges were weighted by the number of common KEGG ortholog groups. Kernel matrices were constructed with a diffusion kernel (Kondor & Lafferty, 2002).

The resulting kernel matrices were centered and normalized. In cross-validation, only the training part of the matrices was preprocessed for learning, while prediction, centering and normalization were performed on the entire data set. The prediction task was defined through the classification matrix of genes and their associated GO slim terms from $\mathbf{R}_{12}$.

### 4.4. Preprocessing for Early Integration

The gene-related data matrices prepared for kernel-based fusion were also used for early integration and were concatenated into a single table. Each row in the table represented a gene profile obtained from all available data sources. For our case study, each gene was characterized by a fixed 9,362-dimensional feature vector. For each GO slim term, we then separately developed a classifier with a random forest of classification trees and reported cross-validated results.

## 5. Results and Discussion

### 5.1. Predictive Performance

Table 1 presents the cross-validated $F_1$ scores the data set of slim terms. The accuracy of our matrix factorization approach is at least comparable to MKL and substantially higher than that of early integration by random forests. The performance of all three fusion approaches improved when more genes and hence more data were included in the study. Adding genes with sparser profiles also increased the overall data sparsity, to which the factorization approach was least sensitive.

The accuracy for nine selected GO terms is given in Table 2. Our factorization approach yields consistently higher $F_1$ scores than the other two approaches. Again, the early integration by random forests is inferior to both intermediate integration methods. Notice that, with one or two exceptions, $F_1$ scores are very high. This is important, as all nine gene processes and functions observed are relevant for current research of *D. discoideum* where the methods for data fusion can yield new candidate genes for focused experimental studies.

Our fusion approach is faster than multiple kernel learning. Factorization required 18 minutes of runtime on a standard desktop computer compared to 77 minutes for MKL to finish one iteration of cross-validation on a whole-genome data set.

*Table 1.* Cross-validated $F_1$ scores for fusion by matrix factorization (MF), a kernel-based method (MKL) and random forests (RF).

| *D. discoideum* task | MF | MKL | RF |
| --- | --- | --- | --- |
| 100 genes | 0.799 | 0.781 | 0.761 |
| 1000 genes | 0.826 | 0.787 | 0.767 |
| Whole genome | 0.831 | 0.800 | 0.782 |

### 5.2. Sensitivity to Inclusion of Data Sources

Inclusion of additional data sources improves the accuracy of prediction models. We illustrate this in Fig. 3(a), where we started with only the target data source $\mathbf{R}_{12}$ and then added either $\mathbf{R}_{13}$ or $\mathbf{\Theta}_1$ or both. Similar effects were observed when we studied other combinations of data sources (not shown here for brevity).

### 5.3. Sensitivity to Inclusion of Constraints

We varied the sparseness of gene constraint matrix $\mathbf{\Theta}_1$ by holding out a random subset of protein-protein interactions. We set the entries of $\mathbf{\Theta}_1$ that correspond to hold-out constraints to zero so that they did not affect the cost function during optimization. Fig. 3(b) shows that including additional information on genes in the form of constraints improves the predictive performance of the factorization model.

### 5.4. Matrix Factor Initialization Study

We studied the effect of initialization by observing the error of the resulting factorization after one and after twenty iterations of factorization matrix updates,

*Table 2.* Gene ontology term-specific cross-validated $F_1$ scores for fusion by matrix factorization (MF), a kernel-based method (MKL) and random forests (RF).

| GO term name | Term identifier | Namespace | Size | MF | MKL | RF |
|---|---|---|---|---|---|---|
| Activation of adenylate cyclase activity | 0007190 | BP | 11 | 0.834 | 0.770 | 0.758 |
| Chemotaxis | 0006935 | BP | 58 | 0.981 | 0.794 | 0.538 |
| Chemotaxis to cAM | 0043327 | BP | 21 | 0.922 | 0.835 | 0.798 |
| Phagocytosis | 0006909 | BP | 33 | 0.956 | 0.892 | 0.789 |
| Response to bacterium | 0009617 | BP | 51 | 0.899 | 0.788 | 0.785 |
| Cell-cell adhesion | 0016337 | BP | 14 | 0.883 | 0.867 | 0.728 |
| Actin binding | 0003779 | MF | 43 | 0.676 | 0.664 | 0.642 |
| Lysozyme activity | 0003796 | MF | 4 | 0.782 | 0.774 | 0.754 |
| Sequence-specific DNA binding TFA | 0003700 | MF | 79 | 0.956 | 0.894 | 0.732 |

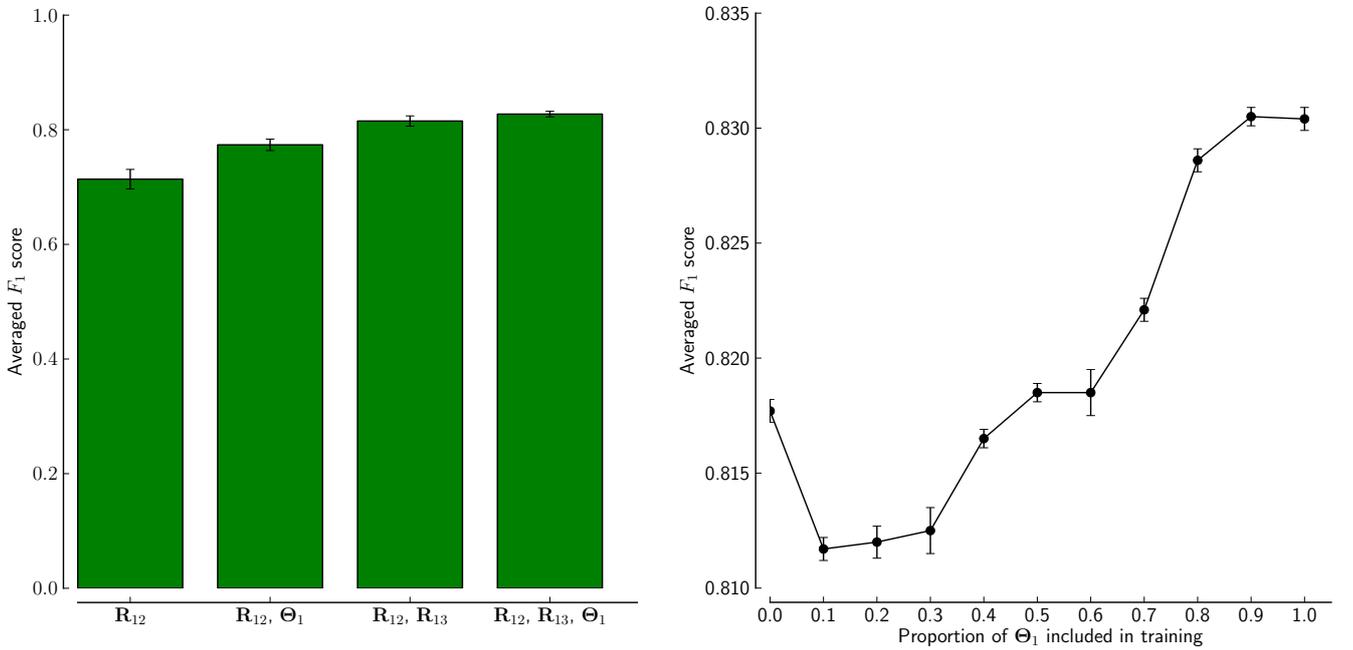

*Figure 3.* Adding new data sources (a) or incorporating more object-type-specific constraints in $\mathbf{\Theta}_1$ (b) both increase the accuracy of the matrix factorization-based models.

the latter being about one fourth of the iterations required for factorization to converge. We estimate the error relative to the optimal $(k_1, k_2, \ldots, k_6)$-rank approximation given by the SVD. For iteration $v$ and matrix $\mathbf{R}_{ij}$ the error is computed by:

$$\mathrm{Err}_{ij}(v) = \frac{||\mathbf{R}_{ij} - \mathbf{G}_i^{(v)} \mathbf{S}_{ij}^{(v)} (\mathbf{G}_j^T)^{(v)}|| - d_F(\mathbf{R}_{ij}, [\mathbf{R}_{ij}]_k)}{d_F(\mathbf{R}_{ij}, [\mathbf{R}_{ij}]_k)},$$
(11)

where $\mathbf{G}_i^{(v)}$, $\mathbf{G}_j^{(v)}$ and $\mathbf{S}_{ij}^{(v)}$ are matrix factors obtained after executing $v$ iterations of factorization algorithm. In Eq. (11) $d_F(\mathbf{R}_{ij}, [\mathbf{R}_{ij}]_k) = ||\mathbf{R}_{ij} - \mathbf{U}_k \mathbf{\Sigma}_k \mathbf{V}_k^T||$ denotes the Frobenius distance between $\mathbf{R}_{ij}$ and its $k$-

rank approximation given by the SVD, where $k = \max(k_i, k_j)$ is the approximation rank. $\mathrm{Err}_{ij}(v)$ is a pessimistic measure of quantitative accuracy because of the choice of $k$. This error measure is similar to the error of the two-factor non-negative matrix factorization from (Albright et al., 2006).

Table 3 shows the results for the experiment with 1000 most GO-annotated *D. discoideum* genes and selected factorization ranks $k_1 = 65$, $k_2 = 35$, $k_3 = 13$, $k_4 = 35$, $k_5 = 30$ and $k_6 = 10$. The informed initialization algorithms surpass the random initialization. Of these, the random Acol algorithm performs best in terms of accuracy and is also one of the simplest.

*Table 3.* Effect of initialization algorithm on predictive power of factorization model.

| Method | Time $\mathbf{G}^{(0)}$ | Storage $\mathbf{G}^{(0)}$ | Err$_{12}$(1) | Err$_{12}$(20) |
|---|---|---|---|---|
| Rand. | 0.0011 s | 618K | 5.11 | 3.61 |
| Rand. C | 0.1027 s | 553K | 2.97 | 1.67 |
| Rand. Acol | 0.0654 s | 505K | 1.59 | 1.30 |
| K-means | 0.4029 s | 562K | 2.47 | 2.20 |
| NNDSVDa | 0.1193 s | 562K | 3.50 | 2.01 |

### 5.5. Early Integration by Matrix Factorization

Our data fusion approach simultaneously factorizes individual blocks of data in $\mathbf{R}$. Alternatively, we could also disregard the data structure, and treat $\mathbf{R}$ as a single data matrix. Such data treatment would transform our data fusion approach to that of early integration and lose the benefits of structured system and source-specific factorization. To prove this experimentally, we considered the 1,000 most GO-annotated *D. discoideum* genes. The resulting cross-validated $F_1$ score for factorization-based early integration was 0.576, compared to 0.826 obtained with our proposed data fusion algorithm. This result is not surprising as neglecting the structure of the system also causes the loss of the structure in matrix factors and the loss of zero blocks in factors $\mathbf{S}$ and $\mathbf{G}$ from Eq. (3). Clearly, data structure carries substantial information and should be retained in the model.

### 6. Conclusion

We have proposed a new data fusion algorithm. The approach is flexible and, in contrast to state-of-the-art kernel-based methods, requires minimal, if any, pre-processing of data. This latter feature, together with its excellent accuracy and time response, are the principal advantages of our new algorithm.

Our approach can model any collection of data sets, each of which can be expressed in a matrix. The gene function prediction task considered in the paper, which has traditionally been regarded as a classification problem (Larranaga et al., 2006), is just one example of the types of problems that can be addressed with our method. We anticipate the utility of factorization-based data fusion in multiple-target learning, association mining, clustering, link prediction or structured output prediction.

### Acknowledgments

We thank Janez Demšar and Gad Shaulsky for their comments on the early version of this manuscript. We acknowledge the support for our work from the Slovenian Research Agency (P2-0209, J2-9699, L2-1112), National Institute of Health (P01-HD39691) and European Commission (Health-F5-2010-242038).